# Towards an Understanding of the Role Operator Limb Dynamics Plays in Haptic Perception of Stiffness

Mohit Singhala[1] and Jeremy D. Brown[2]

*Abstract*— Creating haptic interfaces capable of rendering the rich sensation needed for dexterous manipulation is crucial for the advancement of human-in-the-loop telerobotic systems (HiLTS). One limiting factor has been the absence of detailed knowledge of the effect of operator limb dynamics and haptic exploration dynamics on haptic perception. We propose to begin investigations of these effects with single-joint haptic exploration and feedback of physical and virtual environments. Here, we present our experimental apparatus, a 1-DoF rotational kinesthetic haptic device and electromyography (EMG) system, along with preliminary findings from our efforts to investigate the change in stiffness discrimination thresholds for differing exploration velocities. Result trends indicate a possible relationship between exploration velocity and discrimination thresholds, as well as a complex interaction between muscle activation, exploration velocity, and haptic feedback.

## I. INTRODUCTION

The Human body is capable of highly dexterous manipulation, an ability developed and refined through years of practice performing manipulation tasks in different environments under varying conditions [1]. These manipulations rely on sensory information resulting from our interactions with the environment [2], [3]. The value of haptic cues in guiding this interaction makes it essential to incorporate haptic feedback mechanisms into HiLTS that aim to extend the dexterity of the natural limb to remote and virtual environments. Unfortunately, robust kinesthetic haptic feedback often comes at the expense of control stability. While the majority of HiLTS research focuses on improving teleoperator transparency through control stability, the results are often based on simplified models of the operator's limb dynamics that do not account for their varying nature [4].

Achieving true transparency in HiLTS will require a broadening in focus to include more explicitly the dynamics of the entire system (man and machine) given that both systems are dynamically coupled at the point with which they make contact. In particular, an understanding of the effect of operator limb dynamics on haptic perception is needed. Likewise are investigations into the effect of haptic exploration dynamics on haptic perception. In this manuscript, we begin to explore the role of limb dynamics and exploration dynamics in haptic perception. We present a modular experimental haptic device that allows for single joint exploration of both physical and virtual environments. Using this device, we performed a preliminary psychophysical study to investigate the effects of exploration velocity on the perception of stiffness.

*This work was supported by an NSF Grant# 1657245
[1] Mohit Singhala is with the Department of Mechanical Engineering, Johns Hopkins University, Baltimore, MD, USA mohit.singhala@jhu.edu
[2] Jeremy D. Brown is with the Department of Mechanical Engineering, Johns Hopkins University, Baltimore, MD, USA jdelainebrown@jhu.edu

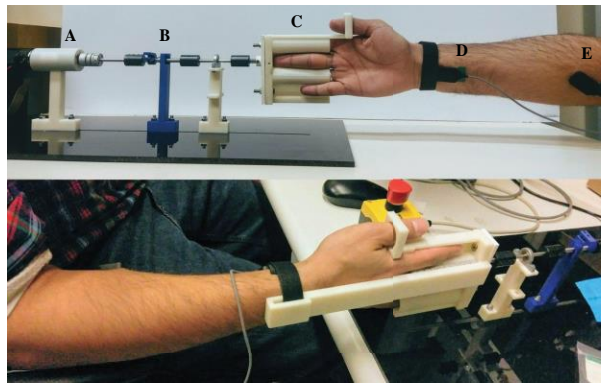

Fig. 1. Top: Experimental setup with A) Housing for the motor, B) Fixture for real torsion springs, C) Custom hand fixture, surface EMG electrodes for D) Pronator Quadratus and E) Pronator Teres muscles. Bottom: Side view of the experimental setup, illustrating the hand fixture.

## II. METHODS

### A. Experimental Setup

The experimental apparatus consists of a custom direct-drive 1-DoF rotary kinesthetic haptic device (see Fig. 1). The device features a Maxon RE30 (15 Watt) motor equipped with a 3-channel Maxon Type L-225787 (1024 CPT) encoder and is driven by a Quanser AMPAQ-L4 Linear current amplifier. Data acquisition and control is provided through a Quanser QPIDe PCI DAQ with a Matlab/Simulink and Quarc real-time software interface. A custom hand fixture, attached via a rotary shaft, allows for a unique alternating finger flexion/extension grip whose purpose is to limit flexion or extension of the whole hand and the influence of other muscles. Velcro bands are used to tie the fixture at the wrist to minimize flexion and extension of the wrist. The elbow is placed at a height adjustable support (not pictured) to align the forearm axis of rotation with the device's rotational axis and to limit radial and ulnar deviation. The setup also includes a removable fixture for real torsion springs, which was disengaged for this set of experiments. A Delsys Bagnoli™ Surface EMG system was used for recording muscle activity. Using Matlab, the EMG data was full-wave rectified, and the DC offset was removed. A linear envelope was then obtained by applying a low-pass 3rd order Butterworth filter with cut-off frequency of 5.5 Hz. Here EMG is used as a proxy for arm dynamics as it has been

shown in prior studies to correlate with dynamic proprieties such as arm impedance [5]

*B. Procedure*

In this preliminary study, we investigated the ability of n=3 participants to distinguish between two virtual springs of different stiffness under two different exploration velocities. All participants were consented according to a protocol approved by the Johns Hopkins School of Medicine Institutional Review Board (Study# IRB00148746). A weighted 1 up/3 down staircase was used in a standard 2AFC same-different task based on the values suggested by Garcia-Perez [6] for a proportion correct target of 83.15%. The staircase was initialized at a value twice that of the reference stimuli and the up stepsize was defined as 10% of the reference stimuli. The ratio of down stepsize and up stepsize was 0.7393, and the staircase terminated after ten reversals. The average of the last eight reversals was used to determine the stiffness-discrimination threshold.

The experiment was performed in a single session consisting of two runs, separated by a five-minute break. Each run involved multiple trials. In each trial, the participant was presented with a virtual linear torsion spring with a spring constant of 1.11 mNm/deg (reference) and another virtual linear torsion spring with a higher stiffness determined by the staircase algorithm. The reference level remained constant throughout the experiment, and the order of presentation for the two springs was chosen randomly. The participants adhered to a strict exploration routine. Each exploration was defined as one rotation of the forearm from normal position (as shown in Fig. 1) to 90 degrees pronation and one rotation back to normal. The participant was provided a console with two buttons labeled "Same" and "Different," which they used to indicate their response. A metronome was used to guide the participants' timing. Each participant performed one run with the metronome set at 45 beats per minute (67.5 deg/s) and one run at 75 beats per minute (112.5 deg/s). The order of exploration velocity was selected randomly for each participant. To ensure repeatability, an LED was used to visually alert the user when they were within 2.5 degrees of their target position. The velocity and the LED trigger were selected based on a series of preliminary experiments to ensure that the trials were comfortable and that the desired exploration was achieved. Each run was preceded by a training session that lasted at least 2 minutes to ensure the participants were able to follow the routine. The experimenter visually monitored each exploration with respect to the metronome. Unsatisfactory trials were repeated before moving ahead on the staircase. EMG measurements were recorded for each trial on the Pronator Quadratus (PQ) and Pronator Teres (PT) muscles.

### III. RESULTS

The results of the stiffness discrimination experiment are listed in Table I. The torque vs. angular displacement profile for the virtual springs at the minimum discrimination threshold for the two angular velocities are shown in Fig. 2

TABLE I
STIFFNESS DISCRIMINATION THRESHOLDS FOR n=3 PARTICIPANTS AT TWO ANGULAR VELOCITIES

| Angular Velocity | P1 | P2 | P3 |
|---|---|---|---|
| 67.5 deg/s | 116% | 96.3% | 100.3% |
| 112.5 deg/s | 95.4% | 83.8% | 88.4% |

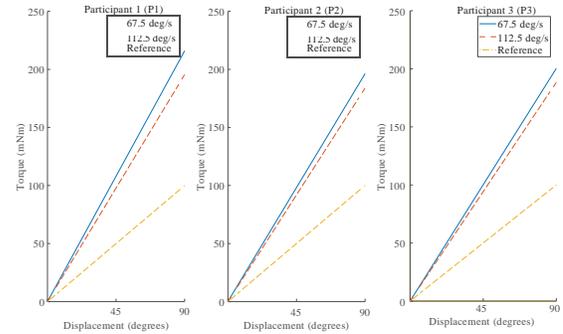

Fig. 2. Torque vs. angular displacement profile representing the stiffness of the reference virtual springs and the virtual springs at the discrimination threshold for 67.5 deg/s and 112.5 deg/s exploration velocities.

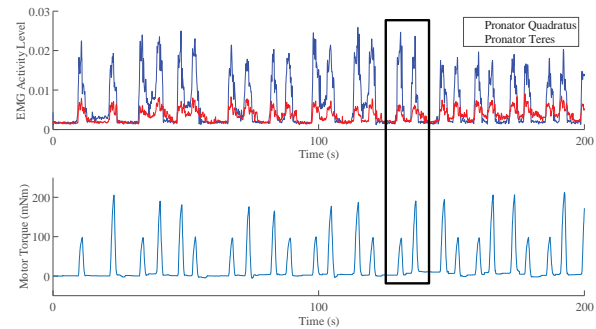

Fig. 3. Sample linear envelope of EMG activity and corresponding motor torque from participant P1's 67.5 deg/s run. Highlighted area (black rectangle) represents a sample case where EMG activity is higher while exploring the reference virtual spring compared to the EMG activity a virtual spring with higher stiffness

for all three participants. All three participants demonstrated a higher discrimination threshold at 67.5 deg/s compared to 112.5 deg/s. In general, higher EMG activity was observed in the PQ muscle compared to the PT muscle. In some trials, the EMG activity for a lower stiffness spring was found to be higher than for a higher stiffness spring (see box in Fig. 3).

### IV. CONCLUSIONS

The average stiffness discrimination threshold for the 1.11 mNm/deg spring was found to be 89.18% at 112.5 deg/s and 105.26% at 67.5 deg/s suggesting that a relationship may exist between exploration velocity and haptic perception. The literature on forearm stiffness discrimination is limited, with most studies reporting constant torque discrimination or stiffness discrimination with a pinch-grasp style of grip [7], [8], [9]. The use of a pinch-grasp makes it difficult to draw comparisons with the results presented here as the mechanics of the motions and muscle activation likely differ significantly. Based on the values reported in literature, the staircase was initialized at 200% the reference stimuli with an expected threshold below 20%. Since our results reflect a higher threshold, we will initialize the staircase at a higher value for future experiments. To enable this, we plan to use a motor capable of simulating virtual springs of higher stiffness values. Trials will be performed with a higher reference stiffness and with a larger sample size to verify the robustness of these findings and to see if Weber′s law is satisfied. Additional experiments are also planned to compare the exploration of real springs to virtual springs at different velocities and at different levels of muscle fatigue, to further analyze the role of human body dynamics in perception. We are also considering the use of signal detection theory for estimation of the sensitivity index to account for response bias of the participants.

While our EMG measures are purely observational at this point, they do suggest that limb dynamics have a potentially complex relationship with limb movement and externally applied forces. It was also anecdotally noticed that the participants had a higher tendency to report the stiffness to be the Same, when the peak of EMG envelope was lower for the higher stiffness compared to the reference. We are working towards establishing a criterion that allows for better statistical interpretation of the EMG data in the context of this task. The inter-subject variability and the nature of surface EMG signals has prevented us from reaching specific conclusions yet. The constraints associated with surface EMG data analysis of deep muscles like Pronator Quadratus will also be considered carefully in our future work. We are currently trying to compare the trends in EMG signals for each trial with the participants error rate, which we believe might be a good indicator of the role that EMG has to play in perception of stiffness. Since the users were asked to repeat an exploration if the desired velocity was not achieved and a manual override was used if the participant hit the wrong button or registered a response multiple times, automated analysis of EMG signals could not be carried out in Matlab. This is a weakness of the experiment protocol that has been fixed to enable robust analysis of EMG data in the future experiments.

## ACKNOWLEDGEMENT

This material is based upon work supported by the National Science Foundation under NSF Grant# 1657245.